\documentclass{article}
\usepackage{ijcai21}

\usepackage{times}
\usepackage{soul}
\usepackage{url}
\usepackage[hidelinks]{hyperref}
\usepackage[utf8]{inputenc}
\usepackage[small]{caption}
\usepackage{graphicx}
\usepackage{amsmath}
\usepackage{amsthm}
\usepackage{booktabs}
\usepackage{algorithm}
\usepackage{algorithmic}
\usepackage{CJKutf8}
\usepackage{makecell}
\usepackage{multirow}
\usepackage{diagbox}
\usepackage{pifont}
\usepackage{amssymb}
\usepackage{colortbl}
\usepackage{arydshln}
\usepackage{mathrsfs}

\newcommand{\sigmoid}{\operatorname{sigmoid}}

\newcommand{\MW}{\boldsymbol{\mathit{W}}} 

\newcommand{\SetV}{\mathbb{V}} 
\newcommand{\SetE}{\mathbb{E}}
\newcommand{\SetR}{\mathbb{R}}
\newcommand{\SetP}{\mathbb{P}}
\newcommand{\SetN}{\mathbb{N}}

\newcommand{\Vh}{\boldsymbol{\mathit{h}}}
\newcommand{\Ve}{\boldsymbol{\mathit{e}}}
\newcommand{\Vs}{\boldsymbol{\mathit{s}}}
\newcommand{\Va}{\boldsymbol{\mathit{a}}}

\newcommand{\Sg}{\mathit{g}}
\newcommand{\Se}{\mathit{e}}
\newcommand{\Sa}{\mathit{a}}

\newcommand{\CalG}{\mathcal{G}} 
\newcommand{\CalL}{\mathcal{L}}
\newcommand{\CalU}{\mathcal{U}}
\newcommand{\CalY}{\mathcal{Y}}

\newcommand{\newcite}[1]{\citeauthor{#1}~\shortcite{#1}}
\newcommand{\citep}[1]{\citeauthor{#1},~\citeyear{#1}}

\usepackage{xspace}
\newcommand{\system}{{\textsc{DdaMS}}\xspace}
\newcommand{\method}{{\textsc{DdaDA}}\xspace}

\usepackage{xcolor, soul}
\definecolor{xcfcolor}{rgb}{0.858, 0.188, 0.478}

\title{Dialogue Discourse-Aware Graph Model and Data Augmentation \\ for Meeting Summarization}

\author{
	Xiachong Feng
	\and
	Xiaocheng Feng 
	\and
	Bing Qin
	\and 
	Xinwei Geng
	\affiliations
	Harbin Institute of Technology, China
	\emails
\{xiachongfeng, xcfeng, bqin, xwgeng\}@ir.hit.edu.cn
}

\begin{document}

\maketitle

\begin{abstract}
Meeting summarization is a challenging task due to its dynamic interaction nature among multiple speakers and lack of sufficient training data.
Existing methods view the meeting as a linear sequence of utterances while ignoring the diverse relations between each utterance. 
Besides, the limited labeled data further hinders the ability of data-hungry neural models.
In this paper, we try to mitigate the above challenges by introducing dialogue-discourse relations.
First, we present a \textbf{D}ialogue \textbf{D}iscourse-\textbf{A}ware \textbf{M}eeting \textbf{S}ummarizer (\system) to explicitly model the interaction between utterances in a meeting by modeling different discourse relations.
The core module is a relational graph encoder, where the utterances and discourse relations are modeled in a graph interaction manner.
Moreover, we devise a \textbf{D}ialogue \textbf{D}iscourse-\textbf{A}ware \textbf{D}ata \textbf{A}ugmentation (\method) strategy to  construct a pseudo-summarization corpus from existing input meetings, which is 20 times larger than the original dataset and can be used to pretrain \system. 
Experimental results on AMI and ICSI meeting datasets show that our full system can achieve SOTA performance Our codes and outputs are available at: \url{https://github.com/xcfcode/DDAMS/}. 
\end{abstract}

\section{Introduction}
\begin{figure}[t]
	\centering
	\includegraphics[scale=0.49]{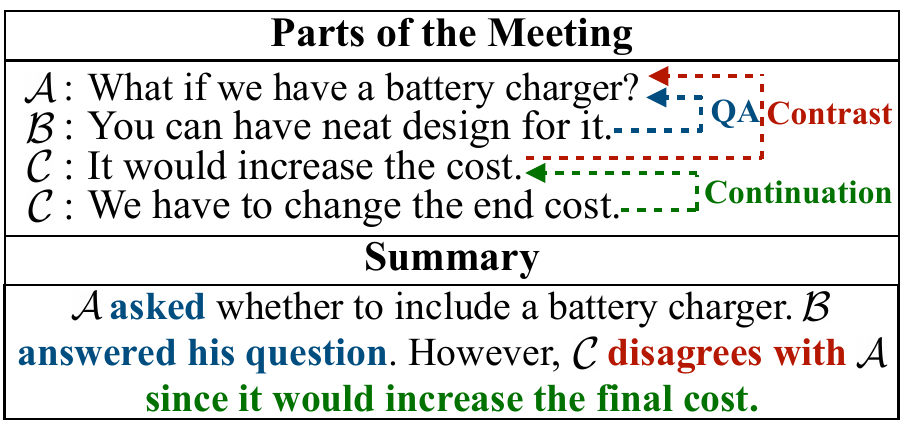}
	\caption{An example of a meeting with its corresponding summary. {\em QA}, {\em Contrast} and {\em Continuation} are dialogue discourse relations, which explicitly show the interaction between utterances.}
	\label{fig:intro}
\end{figure}

Meeting summarization aims to distill the most important information from a recorded meeting into a short textual passage, which could be of great value to people by providing quick access to the essential content of past meetings \cite{meeting-old0}. An example is shown in Figure \ref{fig:intro}, which includes 3 speakers ($\mathcal{A}$, $\mathcal{B}$ and $\mathcal{C}$) and their corresponding utterances, as well as a human-written summary. 

\begin{figure}[t]
	\centering
	\includegraphics[scale=0.4]{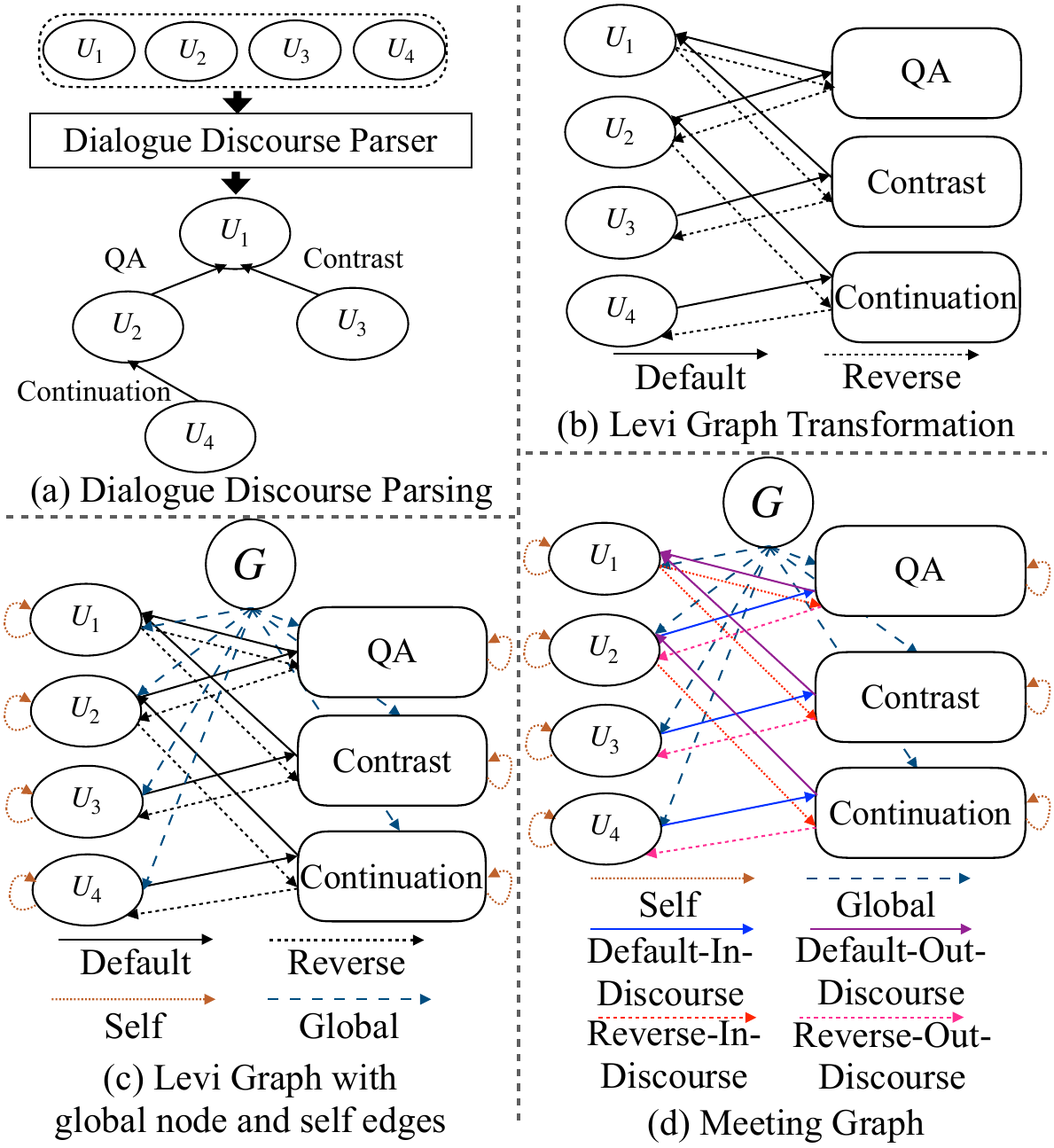}
	\caption{Illustration of meeting graph construction process.}
	\label{fig:graph}
\end{figure}

Recently, neural-based sequence-to-sequence methods have become the mainstream mode for generating meeting summaries via accurately modeling the refined semantic representation of the original meeting text.
\newcite{sentgate} incorporate utterance-level dialogue acts to enhance the representation for each utterance. 
\newcite{acl19li} consider topics as a structure information to enrich the meeting representation.
However, we claim that the current research suffers from two problems.
One is sequential text modeling.
A meeting is a dynamic information exchange flow, which is informal, verbose, and less structured than traditional documents \cite{incoherence}. 
But all previous works adopt a sequential modeling strategy for encoding the meeting, hindering the exploration of inherently rich interactive relations between utterances, which makes the meeting modeling inadequate.
The other one is the lack of sufficient training data.
As we know, the scale of corpus is significant to the performance of neural models, the data size of meeting summarization is only one-thousandth of the traditional news summarization (e.g. 137 meetings in AMI v.s. 312K articles in CNN/DM).

To address the above concerns, we draw support from dialogue discourse, a dialogue-specific structure, which can provide pre-defined relations between utterances \cite{discourse-useful1}. 
As shown in Figure \ref{fig:intro}, {\em QA}, {\em Contrast} and {\em Continuation} are three dialogue discourse relations, which explicitly show the information flow and interactions between utterances. After knowing these relations, we can generate better summaries.
Furthermore, we find that a question often sparks a discussion in a meeting.
As shown in Figure \ref{fig:intro}, the discussion about  ``design" and ``cost" surrounds the salient term ``battery charger" in the question.
Therefore, we assume that a question tends to contain core topics or key concepts that can be used as a pseudo summary for subsequent discussion.

In this paper, we propose a \textbf{D}ialogue \textbf{D}iscourse-\textbf{A}ware \textbf{M}eeting \textbf{S}ummarizer (\system, see \S\ref{sec:model}) for modeling the dialogue discourse relations among utterances. Specifically, we first convert meeting utterances with discourse relations into a meeting graph where utterance vertices interact with relation vertices. Then we design a graph-to-sequence framework to generate meeting summaries.
Besides, to tackle the insufficient training problem, we devise a \textbf{D}ialogue \textbf{D}iscourse-\textbf{A}ware \textbf{D}ata \textbf{A}ugmentation (\method, see \S\ref{sec:dst}) strategy, which constructs a pseudo-summarization corpus from existing input meetings. In detail, we use the {\em QA} discourse relation to identify the question as pseudo-summary, then we view subsequent utterances with associated discourse relations as a pseudo-meeting. Finally, we can construct a pseudo-summarization corpus, which is 20 times larger than the original dataset and can be used to pretrain \system. 

We conduct various experiments on the widely used AMI \cite{ami} and ICSI \cite{icsi} datasets. 
The results show the effectiveness of \system method and \method strategy.
In summary, 
(1) We make the first attempt to successfully explore dialogue discourse to model the utterances interactions for meeting summarization; 
(2) We devise a dialogue discourse-aware data augmentation strategy to alleviate the data insufficiency problem; 
(3) Extensive experiments show that our model achieves SOTA performance.

\section{Preliminaries}
In this section, we describe the task definition and give a brief introduction to dialogue discourse. 

\subsection{Task Definition}
Meeting summarization aims at producing a summary ${\CalY}$ for the input meeting ${\CalU}$, where ${\CalU}$ consists of $|{\CalU}|$ utterances $[u_1,u_2,...u_{|{\CalU}|}]$ and ${\CalY}$ consists of $|{\CalY}|$ words $[y_1,y_2,...y_{|{\CalY}|}]$. The $i$-th utterance of the meeting can be represented as a sequence of words $u_i=[u_{i,1},u_{i,2},...u_{i,|u_i|}]$, where $u_{i,j}$ denotes the $j$-th word of $i$-th utterance. Each utterance $u_i$ associates with a speaker $p_i \in {\SetP}$ , $\SetP$ being a set of participants.

\subsection{Dialogue Discourse}\label{sec:dd}
Dialogue discourse indicates relations between discourse units in a conversation (i.e. utterances in a meeting). This dependency-based structure allows relations between non-adjacent utterances that is applicable for multi-party conversations \cite{Li2021DADgraphAD}. There are 16 discourse relations \cite{stac} in total: {\em comment, clarification-question, elaboration, acknowledgment, continuation, explanation, conditional, QA, alternation, question-elaboration, result, background, narration, correction, parallel, contrast}.

\begin{figure*}[!htb]
	\centering
	\includegraphics[scale=0.52]{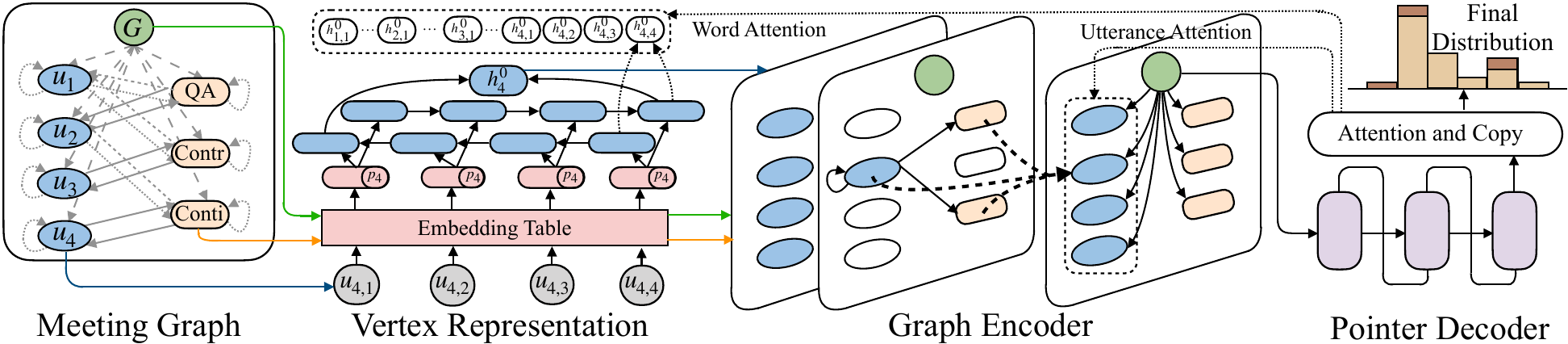}
	\caption{Illustration of our \system model. (1) First, we  construct our meeting graph consisting of three types of vertices: global vertex, utterance vertex and discourse relation vertex. (2) Then, the vertex representation module gives each type of vertex an initial representation. (3) Further, the graph encoder performs convolutional computation over the meeting graph based on the relational graph convolutional network. (4) Finally, the pointer decoder attends to the updated utterance representations and the word representations to generate the summary words either from the fixed-length vocabulary or copy from the input.}
	\label{fig:model}
\end{figure*}

\section{Dialogue Discourse-Aware Meeting Summarizer}\label{sec:model}

In this section, we present our Dialogue Discourse-Aware Meeting Summarizer (\system) that consists of four components: a meeting graph, a vertex representation module, a graph encoder, and a pointer decoder, as shown in Figure \ref{fig:model}.

\subsection{Meeting Graph}\label{sec:dg}

Given meeting utterances, we first use a SOTA dialogue discourse parser \cite{aaai19} to get discourse relations, one relation connects from one utterance to another one with a relation type, as shown in Figure \ref{fig:graph}(a).

Then, we perform the Levi graph transformation, which turns labeled edges into additional vertices \cite{levi2}. Through this transformation, we can model discourse relations explicitly and update the utterance and discourse vertices simultaneously. There are two types of edges in the Levi graph: {\em default} and {\em reverse}.
For example, an edge ($U_3$, {\em Contrast}, $U_1$) in the original graph becomes ($U_3$, {\em default}, {\em Contrast}) and ({\em Contrast}, {\em default}, $U_1$) in the Levi graph, as shown in Figure \ref{fig:graph}(b).

Furthermore, to aggregate non-local information, a global vertex is added that connects all vertices by {\em global} edges and will be used to initialize the decoder. We also add {\em self} edges to aggregate self-information, as shown in Figure \ref{fig:graph}(c).

Note that different types of vertices may have different features that fall in different space \cite{amr-beck}. Specifically, previous works ignore the type of source and target vertices and use the same type of edge to pass information, which may reduce the effectiveness of discourse information. 
Thus, we propose our meeting graph which transforms the {\em default} edge into {\em default-in-discourse} and {\em default-out-discourse} edges and the {\em reverse} edges into {\em reverse-in-discourse} and {\em reverse-out-discourse} edges, as shown in Figure \ref{fig:graph}(d).

Let ${\CalG}_M\!=\!({\SetV}_M, {\SetE}_M, {\SetR}_M)$ denote a meeting graph, with vertices $v_i \!\in\! {\SetV}_M$ and labeled edges $(v_i,r,v_j)\!\in\! {\SetE}_M$, where $r \in {\SetR}_M$ is the relation type of the edge. For our meeting graph, there are six types of relations, where ${\SetR}_{M}$ becomes {\em default-in-discourse, default-out-discourse, reverse-in-discourse, reverse-out-discourse, global, self}.

\subsection{Vertex Representation}
Vertex representation module aims to obtain initial representations for three types of vertices: global vertex, relation vertex and utterance vertex.
For global vertex and relation vertices, we get the initial representation ${\Vh}_i^0$ by looking up from an embedding table. For utterance vertices, we employ a BiLSTM\footnote{We have also tried Transformer \cite{vaswani2017attention} as our backbone model. However, it has too many parameters which can easily over-fit on the small meeting dataset.} as the utterance encoder that encodes the utterance forwardly and backwardly \cite{qin-etal-2020-agif}. As $\overrightarrow{{\Vh}_{i,j}}\!=\!\operatorname{LSTM}_f \left(\overrightarrow{{\Vh}_{i, j-1}}, {\Ve}_{i, j}\right)$, $\overleftarrow{{\Vh}_{i,j}}\!=\!\operatorname{LSTM}_b \left(\overleftarrow{{\Vh}_{i, j+1}}, {\Ve}_{i, j}\right)$, ${\Vh}_{i,j}\!=\![\overrightarrow{{\Vh}_{i,j}};\overleftarrow{{\Vh}_{i,j}}]$, where ${\Vh}_{i,j}$ and ${\Ve}_{i,j}$ denote the hidden state and embedding of word $u_{i,j}$. $;$ denotes concatenation. To involve speaker information, we encode speaker $p_i$ using a one-hot vector and get ${\Ve}_{i,j}$ by concatenating word embedding and one-hot speaker embedding \cite{aaai20}. The representation ${\Vh}_i^0\!=\![\overrightarrow{{\Vh}_{i,|u_i|}};\overleftarrow{{\Vh}_{i,1}}]$ is used as input to the graph encoder.

\subsection{Graph Encoder}\label{sec:ge}
After getting the initial feature ${\Vh}_i^0$ of each vertex $v_i \in {\SetV}_M$, we feed them into the graph encoder to digest the structural information. We use relational graph convolutional networks \cite{rgcn} to capture high-level hidden features considering different types of edges. The convolutional computation for $v_i$ at the $\left(l+1\right)$-th layer takes the representation ${\Vh}^{\left(l\right)}$ at the $l$-th layer as input can be defined as:
\begin{equation}
    {\Vh}_{i}^{(l+1)}= {\textsc{ReLU}} \left(\sum_{r \in {\SetR}_{M}} \sum_{v_{j} \in {\SetN}_{i}^{r}} \frac{1}{\left|{\SetN}_{i}^{r}\right|} {\MW}_{r}^{(l)} {\Vh}_{j}^{(l)}\right)
\end{equation}
where ${\SetN}_{i}^{r}$ denotes the set of neighbors of vertex $v_i$ under relation $r$ and ${\MW}_{r}^{(l)}$ denotes relation-specific learnable parameters at the $l$-th layer.

However, uniformly accepting information from different discourse relations is not suitable for identifying important discourse. Thus, we use the gate mechanism \cite{gate} to control the information passing:
\begin{equation}
    {\Sg}_{j}^{(l)}={\sigmoid} \left({\MW}_{g,r}^{(l)} {\Vh}_{j}^{(l)} \right)
\end{equation}
where ${\MW}_{g,r}^{(l)}$ denotes a learnable parameter under relation type $r$ at the $l$-th layer.
Equipped with the gate mechanism, the convolutional computation can be defined as:
\begin{equation}
    {\Vh}_{i}^{(l+1)}={\textsc{ReLU}} \!\left(\!\sum_{r \in {\SetR}_{M}} \sum_{v_j \in {\SetN}_{i}^{r}} {\Sg}_{j}^{(l)} \frac{1}{\left|{\SetN}_{i}^{r}\right|} {\MW}_{r}^{(l)} {\Vh}_{j}^{(l)}\!\right)\!
\end{equation}

\subsection{Pointer Decoder}
We use a standard LSTM decoder with attention and copy mechanism to generate the summary \cite{bahdanau2014neural,pgn}. The global representation is used to initialize the decoder. At each step $t$, the decoder receives the word embedding of previous word and has decoder state ${\Vs}_t$. The attention distribution is calculated as in \cite{luong}. We consider both word-level and utterance-level attention. The word-level context vector ${\Vh}_t^{wl}$ is computed as:
\begin{equation}
\begin{split}
& {\Se}_{i,j}^{t} = {\Vs}_t^{\top} {\MW}_{a} {\Vh}_{i,j}^0 \\
& {\Va}^{t} =\operatorname{softmax}({\Ve}^{t}) \\
& {\Vh}_t^{wl}=\sum_{i}\sum_{j} {\Sa}_{i,j}^{t} {\Vh}_{i,j}^0 \\
\end{split}
\end{equation}
where ${\MW}_a$ is a learnable parameter and ${\Vh}_{i,j}^0$ is obtained from utterance encoder for $u_{i,j}$. The utterance-level context vector ${\Vh}_t^{ul}$ is calculated similarly to the word-level context vector, except that we use the final outputs of the graph encoder ${\Vh}_i^{(l)}$ which represent utterances to calculate the attention distribution. The final context vector is the concatenation of word-level and utterance-level context vector ${\Vh}_t^{*}=[{\Vh}_t^{wl};{\Vh}_t^{ul}]$, which is then used to calculate generation probability and the final probability distribution \cite{pgn}.

\subsection{Training Objective}
We use maximum likelihood estimation to train our model. Given the ground truth summary ${\CalY}^{*}=[y^{*}_1,y^{*}_2,...,y^{*}_{|{\CalY}^{*}|}]$ for an input meeting ${\CalU}$. We minimize the negative log-likelihood of the target words sequence:
\begin{equation}
{\CalL}=-\sum_{t=1}^{|{\CalY}^{*}|} \log p\left(y_{t}^{*} | y_{1}^{*} \ldots y_{t-1}^{*}, {\CalU}\right)
\end{equation}

\section{Dialogue Discourse-Aware Data Augmentation}\label{sec:dst}
In this section, we introduce Dialogue Discourse-Aware Data Augmentation strategy (\method), which constructs a pseudo-summarization corpus from the original input meetings based on the {\em QA} discourse relation.

\subsection{Pseudo-summarization Corpus Construction}
\begin{figure}[t]
	\centering
	\includegraphics[scale=0.35]{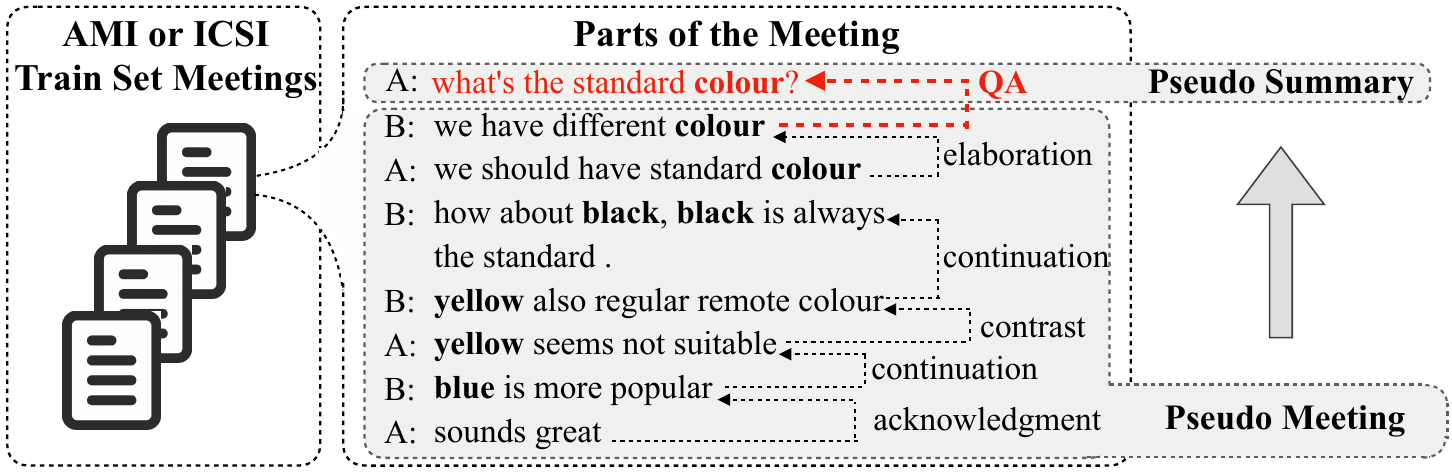}
	\caption{Illustration of how to construct a pseudo meeting-summary pair. Given a meeting from the original meeting train set, we use {\em QA} discourse relation to identify the question in the meeting. Then, the subsequent discussion with discourse relations becomes a pseudo meeting and the question becomes a pseudo summary.}\label{fig:pseudo}
\end{figure}

Given a meeting and its corresponding discourse relations, we find a question often sparks a discussion and contains salient terms or concepts expressed in the discussion. 
As shown in Figure \ref{fig:pseudo}, speaker A asked ``What's the standard colour?", other participants start to discuss this small topic. Thus, we can view the discussion as a pseudo meeting and the question as a pseudo summary of this pseudo meeting\footnote{In our preliminary experiments, we randomly sample 100 pseudo-summarization instances and observe that 68\% of questions can cover important information, such as salient terms and concepts.}. 

According to the above observation, we collect pseudo-summarization data from the original meeting summarization dataset where the question identified by {\em QA} discourse serves as the pseudo summary and $N$ utterances after the question with their associated discourse relations serve as the pseudo meeting\footnote{We conduct experiments with $N \in [6:12]$. Finally, we choose $N$=10 which performs the best.}. Note that there are some uninformative and normal questions such as ``what is this here", which are not suitable for pseudo-summarization corpus construction. Thus, we filter out questions that contain no noun and adjective to make the pseudo data more realistic. Table \ref{tab:pseudo_datasets} shows the statistics for the pseudo-summarization corpus.

\begin{table}[htb]
\small
\centering
        \begin{tabular}{lcc}
            \toprule  
             & \makecell{\textbf{AMI} \\ \textbf{Pseudo Corpus}} & \makecell{\textbf{ICSI} \\ \textbf{Pseudo Corpus}}   \\
            \midrule  
            \# of Original Data &97 &53 \\
            \# of Pseudo Data &1539 &1877   \\
            Avg.Tokens &124.44  &107.44\\
            Avg.Sum  &13.18 &11.97  \\
            \bottomrule 
        \end{tabular}
\caption{Pseudo-summarization corpus statistics. ``\# of Original Data" means the number of original meetings in the train set, ``\# of Pseudo Data" means the number of pseudo meeting-summary pairs, ``Avg.Tokens"  means the average length of pseudo meetings and ``Avg.Sum" means the average length of pseudo summaries.}
\label{tab:pseudo_datasets}
\end{table}

\subsection{Pretraining}
Given the pseudo corpus, we first pretrain \system through a question generation pretraining objective, which pretrains \system to generate questions conditioning on subsequent discussions. Afterward, we fine-tune \system on the meeting summarization dataset. \newcite{qgpre} also reveal that treating question generation as the pretraining objective can help downstream tasks in identifying important contents in the open domain. Our motivations are two-fold: (1) we can potentially augment the training data; (2) since the pseudo data is constructed from the meeting summarization dataset (in-domain), the pre-training can give the model a warm start.

\begin{table*}[t]
\small
\centering
        \begin{tabular}{clcccccc}
            \toprule
            & &  \multicolumn{3}{c}{\textbf{AMI}} & \multicolumn{3}{c}{\textbf{ICSI}} \\
            & \textbf{Model}  & \textbf{R-1} & \textbf{R-2} & \textbf{R-L} & \textbf{R-1} & \textbf{R-2} & \textbf{R-L}  \\
            \midrule
            \multirow{2}{*}{Extractive} & TextRank \cite{textrank}   &35.19 &6.13 &15.70 & 30.72 & 4.69 & 12.97 \\
            & SummaRunner \cite{SummaRuNNer}  &30.98 &5.54 &13.91 & 27.60 & 3.70 & 12.52 \\
            \midrule  
            \multirow{6}{*}{Abstractive} & UNS \cite{emnlp18unsupervised}   & 37.86& 7.84& 13.72 & 31.73 & 5.14 & 14.50 \\
            & Pointer-Generator \cite{pgn}  & 42.60& 14.01& 22.62 & 35.89 & 6.92 & 15.67 \\
            & HRED \cite{HRED}  &49.75 &18.36 &23.90 & 39.15 & 7.86 & 16.25 \\
            & Sentence-Gated \cite{sentgate}  &49.29 &19.31 &24.82 & 39.37 & 9.57 & 17.17 \\
            & TopicSeg \cite{acl19li}  & 51.53& 12.23 & 25.47 & - & - & - \\
            & HMNet \cite{hmnet-new} &52.36 &18.63 &24.00 &\textbf{45.97} &10.14 & 18.54\\
            \midrule 
            \multirow{3}{*}{Ours}  & \system  & 51.42 &20.99 &24.89 & 39.66 & 10.09 & 17.53 \\
            & \system+ \method &\textbf{53.15} &\textbf{22.32} &\textbf{25.67} & 40.41 & \textbf{11.02} & \textbf{19.18} \\
            & \system+ \method(w/o fine-tune) &28.35 &4.67 & 14.92 & 25.94 & 4.18 & 13.92 \\
            \bottomrule 
        \end{tabular}
\caption{Test set results on AMI and ICSI Datasets, where  ``R-1'' is short for ``ROUGE-1'', ``R-2'' for ``ROUGE-2'', ``R-L'' for ``ROUGE-L''. The \system represents the model that is trained only on the meeting dataset. The \system+\method means the model that is pre-trained using pseudo-summarization data and then fine-tuned on meeting dataset. \system+\method(w/o fine-tune) means the model that is pre-trained using pseudo-summarization data and without fine-tuning on the meeting dataset.{\protect\footnotemark[7]}} \label{tab:main_results}
\end{table*}

\section{Experiments}
\noindent \textbf{Datasets}
We experiment on AMI \cite{ami} and ICSI \cite{icsi} datasets. We preprocess the dataset into train, valid and test sets for AMI (97/20/20) and ICSI (53/25/6) separately following \newcite{emnlp18unsupervised}. 
We get discourse relations for one meeting based on the SOTA dialogue discourse parser, namely Deep Sequential \cite{aaai19}, which is trained on the STAC corpus \cite{stac}.\footnote{More information about statistics for datasets, implementation details and relation distributions are shown in the supplementary file.}

\noindent \textbf{Evaluation Metrics}
We adopt ROUGE \cite{rouge} for evaluation and obtain the $F_1$ scores for ROUGE-1, ROUGE-2, and ROUGE-L between the ground-truth and the generated summary respectively.

\noindent \textbf{Baseline Models}
\textbf{TextRank} \cite{textrank} is a graph-based extractive method. \textbf{SummaRunner} \cite{SummaRuNNer} is an extractive method based on hierarchical RNN network. \textbf{UNS} \cite{emnlp18unsupervised} is an unsupervised abstractive method that combines several graph-based approaches. \textbf{Pointer-Generator} \cite{pgn} is an abstractive method equips with copy mechanism. \textbf{HRED} \cite{HRED} is a hierarchical Seq2Seq model. \textbf{Sentence-Gated} \cite{sentgate} is an abstractive method that incorporates dialogue acts by the sentence-gated mechanism. \textbf{TopicSeg} \cite{acl19li} is an abstractive method using a hierarchical attention mechanism at three levels (topic, utterance, word).\footnote{\cite{acl19li} also proposed TopicSeg+VFOA by incorporating vision features in a multi-modal setting.  In this paper, we compare our model with baselines using only textual features.} \textbf{HMNet} \cite{hmnet-new} is an abstractive method that employs a hierarchical model and incorporates part-of-speech and entity information, which is also pre-trained on large-scale news summarization dataset\footnote{\newcite{hmnet-new} directly use news summarization dataset to pre-train their model, which can not be applied to our model, since our \system needs dialogue discourse information.}.

\subsection{Automatic Evaluation}\label{sec:auto_result}
As shown in Table \ref{tab:main_results}, our model \system outperforms various baselines, which shows the effectiveness of dialogue discourse. By pre-training on pseudo-summarization data, our model \system+\method can further boost the performance by a large margin, which shows the need for \method. Specially, \system+\method(w/o fine-tune) still achieves a basic effect, which appears  \system+\method(w/o fine-tune) can simulate as a summarization model in terms of selecting important information.

\subsection{Human Evaluation}

To further assess the quality of the generated summaries, we conduct a human evaluation study.  We choose two metrics: \textbf{relevance} (consistent with the original input) and \textbf{informativeness} (preserves the meaning expressed in the ground-truth). We hired five graduates to perform the human evaluation. They were asked to rate each summary on a scale of 1 (worst) to 5 (best) for each metric. The results are shown in Table \ref{tab:he}.

\begin{table}[!htb]
\small
\centering
        \begin{tabular}{clcc}
            \toprule
            &\textbf{Model} & \textbf{Relevance} & \textbf{Informativeness}  \\
            \midrule 
            \multirow{6}{*}{\rotatebox{90}{AMI}}&Ground-truth  &4.60 &4.56  \\
            \cmidrule{2-4}
            &Sentence-Gated    &3.16 &3.60  \\
            &HMNet    &3.60 &3.72 \\
            &\system  &3.80 &3.76  \\
            &\system+\method  &\textbf{3.84} &\textbf{3.88}  \\
            \midrule  
            \multirow{6}{*}{\rotatebox{90}{ICSI}}&Ground-truth  &4.76 &4.48  \\
            \cmidrule{2-4}
            &Sentence-Gated    &3.32 &3.48  \\
            &HMNet    &3.80 &3.52 \\
            &\system  &3.76  &3.28   \\
            &\system+\method  &\textbf{3.84} &\textbf{3.60}  \\
            \bottomrule 
        \end{tabular}
\caption{Human evaluation results.}
\label{tab:he}
\end{table}
 	 	 
\footnotetext[7]{ROUGE-2 of TopicSeg is significantly lower than other baselines, which contains some anomalies. we reproduce the results of UNS and HMNet by the official codes and evaluate them by PyRouge package, the results are consistent with the original paper, which shows the increase of our ROUGE-2 is reasonable.}

We can see that our \system+ \method achieves higher scores in both relevance and informativeness than other baselines. Ground truth obtains the highest scores compare with generated summaries indicating the challenge of this task.

\subsection{Analysis}
\begin{figure}[htb]
	\centering
	\includegraphics[scale=0.33]{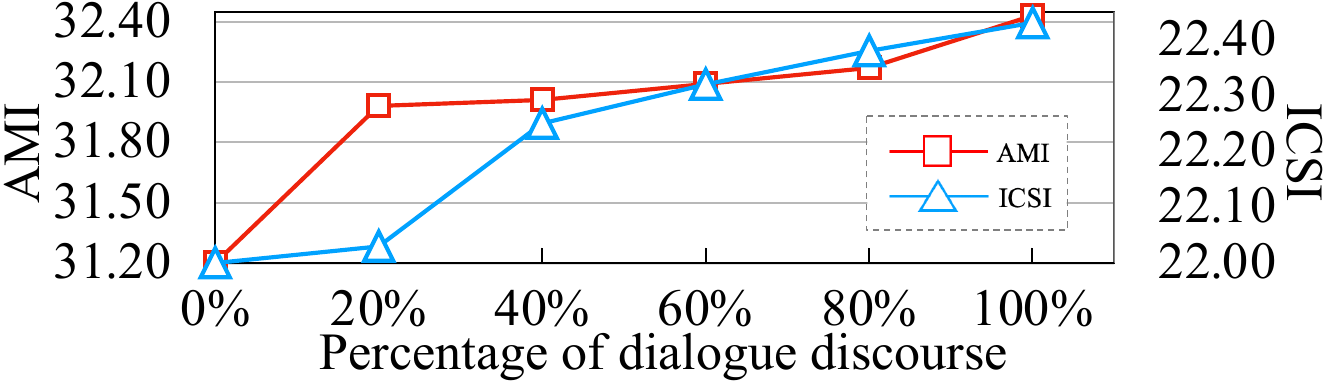}
	\caption{Average ROUGE scores with respect to the number of dialogue discourse relations given at test process.}
	\label{fig:percentage1}
\end{figure}

\paragraph{Effect of the number of dialogue discourse.} We conduct experiments by randomly providing parts of discourse relations to our model \system in the test process. The results are shown in Figure \ref{fig:percentage1}. We can see that the more discourse relations, the higher the ROUGE score, which indicates discourse can do good to summary generation. When given no discourse information, our model gets similar scores compared with HRED, which models utterances sequentially.

\paragraph{Effect of the quality of dialogue discourse.} We train our \system by using discourse relations that are provided by parsers of different qualities\footnotemark[8]. We can see that the higher the quality of discourse parser, the higher the ROUGE score, which potentially indicates high-quality discourse relations can further improve our model.

\footnotetext[8]{We choose parsers of different qualities based on the link prediction $F_1$ score on the STAC corpus test set.}

\begin{figure}[htb]
	\centering
	\includegraphics[scale=0.32]{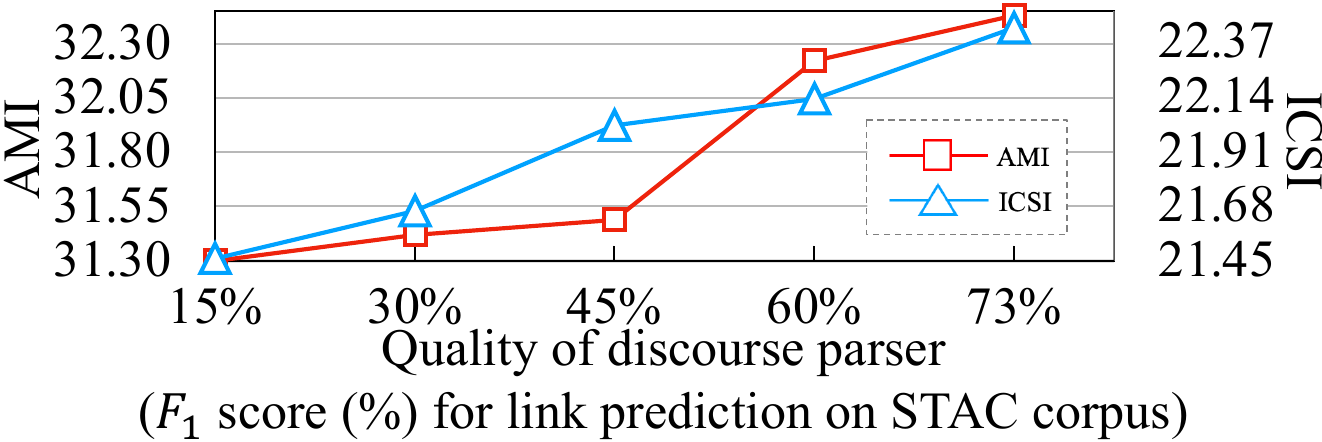}
	\caption{Average ROUGE scores with respect to the quality of discourse parser.}
	\label{fig:percentage2}
\end{figure}

\begin{figure}[t]
	\centering
	\includegraphics[scale=0.25]{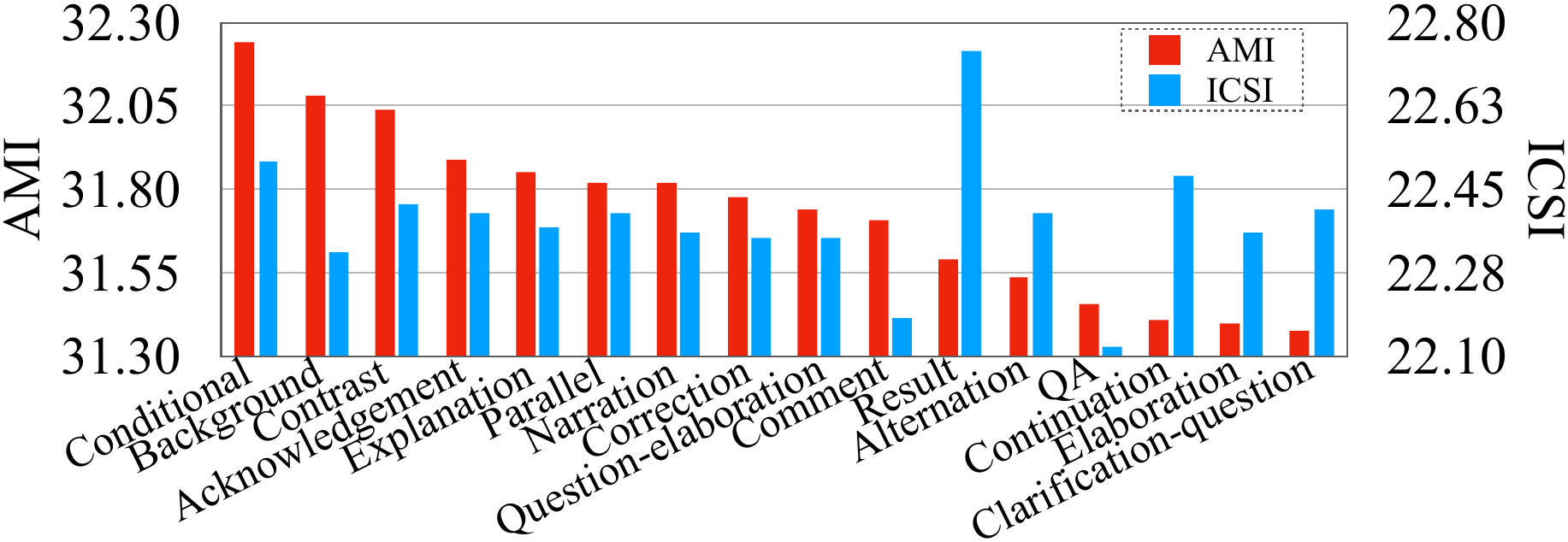}
	\caption{Average ROUGE scores with respect to different types of discourse relations given at test process.}
	\label{fig:types}
\end{figure}

\paragraph{Effect of the type of dialogue discourse.}
To verify the importance of each discourse relation type, we then test our model by giving different discourse relations. The results are shown in Figure \ref{fig:types}. On AMI dataset, the participants take a remote control design project. We can see that {\em Conditional} and {\em Background} provide more accurate and adequate information, facilitating summary generation. On ICSI dataset, meetings focus on academic discussion. We can see that the relation {\em Result} is more important than other relations, we attribute this to the fact that this relation can strongly indicate the outcome of a discussion that covers salient information.

\paragraph{Effect of meeting graph.}
To verify the effectiveness of designing different types of edges for different types of vertices, we replace our meeting graph with Levi graph (shown in Figure \ref{fig:graph}(c)), namely \system (w/ Levi graph). The results are shown in Table \ref{tab:levi_test}. We can see that taking the type of vertices into consideration, our model \system can get better results.

\begin{table}[htb]
\small
\centering
        \begin{tabular}{clccc}
            \toprule  
            & \textbf{Model} & \textbf{R-1} & \textbf{R-2} & \textbf{R-L}  \\
            \midrule  
            \multirow{2}{*}{\rotatebox{90}{AMI}} & \system  &51.42 &\textbf{20.99} &\textbf{24.89} \\
            & \system (w/ Levi graph) &\textbf{51.46} &20.75 &24.31  \\
            \midrule  
            \multirow{2}{*}{\rotatebox{90}{ICSI}} & \system  &\textbf{39.66} &\textbf{10.09} &\textbf{17.53}  \\
            & \system (w/ Levi graph) &39.20 &9.54 &17.48  \\
            \bottomrule 
        \end{tabular}
\caption{Test set results of using meeting graph and Levi graph.}
\label{tab:levi_test}
\end{table}

\paragraph{Effect of pseudo-summarization data.}
To further study the effectiveness of pseudo-summarization data, instead of using questions identified by dialogue discourse as pseudo summaries, we extract questions following two rules: (1) utterances that begin with WH-words (e.g. “what”), (2) utterances that end with a question mark.  We call this Rule-based Data Augmentation (\textsc{RbDa}). We first pre-train our \system on the two types of pseudo data, which are constructed based on \method and \textsc{RbDa} separately. And then fine-tune it on the meeting dataset. The results are shown in Table \ref{tab:pseudo}. We can see that \method is better than \textsc{RbDa}, demonstrating the effectiveness of dialogue discourse. Besides, pretraining on pseudo-summarization data constructed based on \textsc{RbDa} still achieves a better result, which indicates the rationality of our pretraining strategy.

\begin{table}[htb]
\small
\centering
        \begin{tabular}{clccc}
            \toprule  
            & \textbf{Model} & \textbf{R-1} & \textbf{R-2} & \textbf{R-L}  \\
            \midrule  
            \multirow{3}{*}{\rotatebox{90}{AMI}} & \system  &51.42 &20.99 &24.89  \\
            & + \textsc{RbDa} &52.94 &21.96 &25.05  \\
            & + \method &\textbf{53.15} &\textbf{22.32} &\textbf{25.67}  \\
            \midrule  
            \multirow{3}{*}{\rotatebox{90}{ICSI}} & \system  &39.66 &10.09 &17.53  \\
            & + \textsc{RbDa} &39.42 &10.60 &18.19  \\
            & + \method &\textbf{40.41} &\textbf{11.02} &\textbf{19.18}  \\
            \bottomrule 
        \end{tabular}
\caption{The results of pre-training \system on different types of pseudo-summarization data.}
\label{tab:pseudo}
\end{table}

\paragraph{Others.} We show more analyses in the supplementary file, including: (1) utterance-level and word-level attention mechanisms; (2) filtering out useless relations according to Figure \ref{fig:types};  (3) the case study.

\section{Related Work}
\noindent \textbf{Meeting Summarization}
With the abundance of automatic meeting transcripts, meeting summarization attracts more and more attention \cite{murray2010generating,emnlp18unsupervised,meetseg}.
There are two main paradigms in meeting summarization.
In the first paradigm, some works try to incorporate auxiliary information for better modeling meetings.
\newcite{sentgate} incorporated dialogue acts.
\newcite{acl19li} incorporated topics and vision features.
\newcite{crf-discourse} reconstructed the dialogue into a sequence of words based on discourse labels.
\newcite{coling20-domain} investigated the impact of domain
terminologies.
In the second paradigm, some works used news summarization data to alleviate the insufficient training problem.
\newcite{hmnet-new} used news data to pretrain their model. 
\newcite{crf-discourse} used a news summarizer to generate meeting summaries in a zero-shot manner.
In this paper, we propose to use dialogue discourse relations as the auxiliary information to model the interaction between utterances and construct a pseudo-summarization corpus, addressing the data insufficiency problem.

\noindent \textbf{Discourse-Aware News Summarization}
Some works applied a constituency-based discourse structure, namely Rhetorical Structure Theory \cite{RST} to news summarization \cite{discourse-news1,discourse-news2}. These works aimed to analyze the core part of a given sentence and extracted sub-sentential discourse units to form the summary. In this paper, we introduce the dependency-based structure, namely dialogue discourse, for exploring the diverse interactive relations between utterances.

\section{Conclusion}
In this paper, we apply the dialogue discourse to model the diverse interactions between utterances and present a Dialogue Discourse-Aware Meeting Summarizer (\system). 
We design a meeting graph to facilitate the information flow and demonstrate that incorporating dialogue discourse is effective for this task. 
Moreover, we propose a Dialogue Discourse-Aware Data Augmentation (\method) strategy to alleviate the insufficient training problem.
We build a pseudo-summarization corpus by utilizing the {\em QA} discourse relation.
Experiments on AMI and ICSI datasets show that our model achieves new SOTA performances.

\section*{Acknowledgements}
This work is supported by the National Key R\&D Program of China via grant 2018YFB1005103 and National Natural Science Foundation of China (NSFC) via grant 61906053 and 61976073. We thank all the anonymous reviewers for their insightful comments. We also thank Libo Qin, Yibo Sun and Jiaqi Li for helpful discussion. 

\appendix

\section{Experimental Details}
\paragraph{\textbf{Implementation Details}}
For our \system, the dimension of hidden states is set to 200. The embedding size is 300. We use Adam with learning rate of 0.001. Dropout rate is set to 0.5. In test process, beam size is set to 10. For pre-training, we stop training until the model converges on pseudo data. For discourse parser training, we use default parameters and set vocabulary size to 2,500{\protect\footnotemark[9]}. 

\footnotetext[9]{https://github.com/shizhouxing/DialogueDiscourseParsing}

\paragraph{\textbf{Baseline Codes}}
Official codes for each baseline model:
\begin{itemize}
\item \textbf{SummaRunner}: https://github.com/kedz/nnsum
\item \textbf{UNS}: https://bitbucket.org/dascim/acl2018\_abssumm
\item \textbf{Pointer-Generator}: https://github.com/OpenNMT
\item \textbf{Sentence-Gated}: https://github.com/MiuLab/DialSum
\item \textbf{HMNet}: https://github.com/microsoft/HMNet
\end{itemize}

\section{Details of AMI and ICSI Corpus}
Statistics for AMI \cite{ami} and ICSI \cite{icsi} are shown in Table \ref{tab:datasets}. ``\#" means the number of meetings in the dataset, ``Avg.Turns" means the average turns of all meetings, ``Avg.Tokens" means the average length of meetings and ``Avg.Sum" means the average length of summaries.

\begin{table}[htb]
\centering
        \begin{tabular}{lcc}
            \toprule  
             & \makecell{\textbf{AMI}} & \makecell{\textbf{ICSI}}   \\
            \midrule  
            \#  &137 &59 \\
            Avg.Turns &289 &464 \\
            Avg.Tokens &4,757  &10,189  \\
            Avg.Sum  &322 &534  \\
            \bottomrule 
        \end{tabular}
\caption{Statistics for AMI and ICSI datasets.}
\label{tab:datasets}
\end{table}

\section{Distribution of Discourse Relations}
\begin{figure*}[]
	\centering
	\includegraphics[scale=0.5]{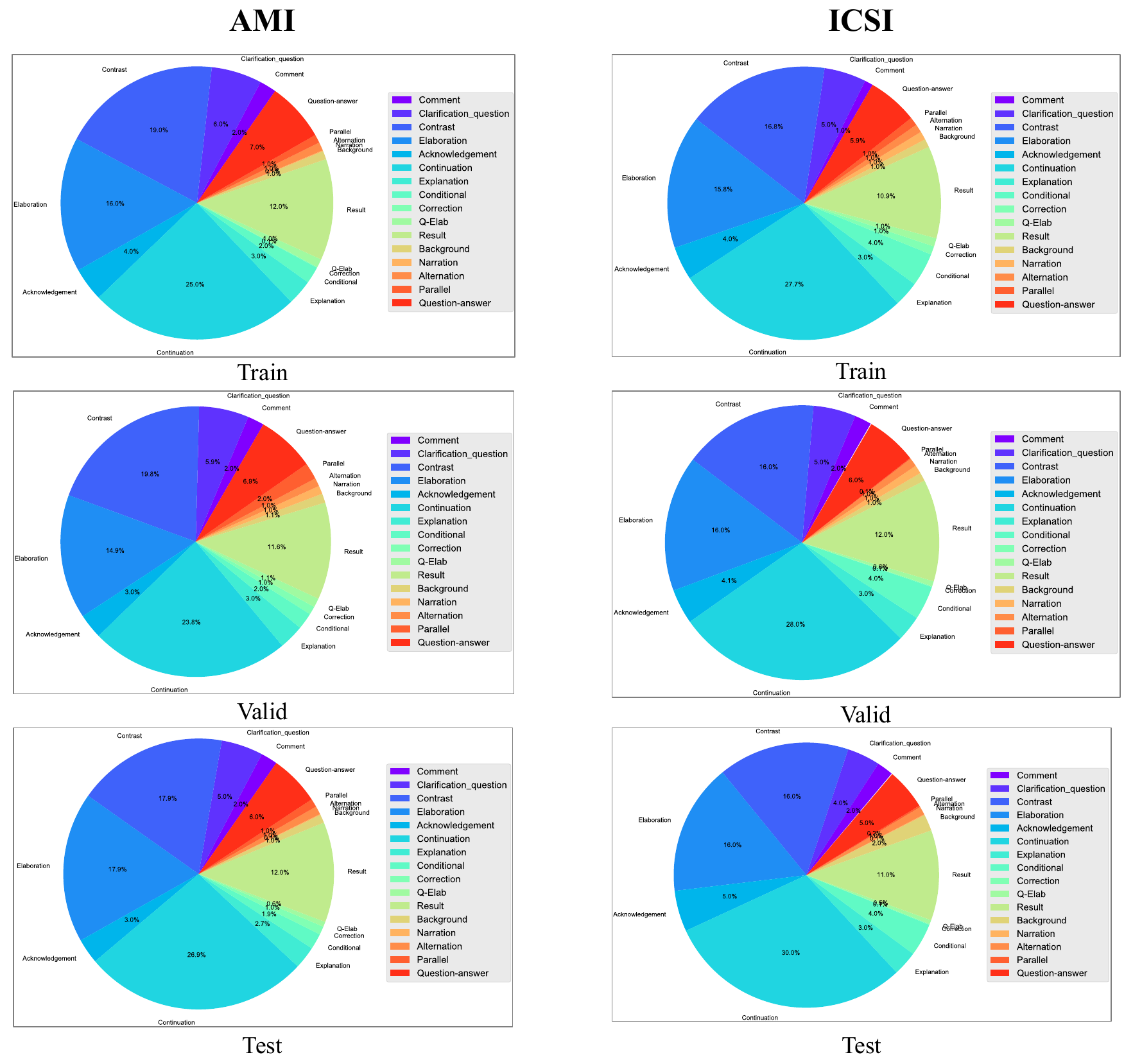}
	\caption{Relation distribution statistics.}
	\label{fig:dis}
\end{figure*}

We get dialogue discourse relations for one meeting based on Deep Sequential \cite{aaai19}, a SOTA dialogue discourse parser which is trained on the STAC corpus \cite{stac}. The final relations distributions for AMI and ICSI are shown in Figure \ref{fig:dis}.

\section{Effect of attention mechanisms}
We conduct ablation studies to show the effectiveness of different types of attention mechanism. The results are shown in Table \ref{tab:attention}.
We can see that word attention is more important.
However, our model achieves the best performance by equipping both word and utterance attentions.
The results reveal the importance of combining both attention mechanisms for meeting summarization task.

\begin{table}[]
\centering
        \begin{tabular}{c|lccc}
            \toprule  
            & \textbf{Model} & \textbf{R-1} & \textbf{R-2} & \textbf{R-L}  \\
            \midrule  
            \multirow{3}{*}{\rotatebox{90}{AMI}} & \system  &\textbf{51.42} &\textbf{20.99} &24.89  \\
            &  \ \ w/o \textit{utter-attn} &51.22 &20.57 &\textbf{25.02}  \\
            &  \ \ w/o \textit{word-attn} &50.27 &19.81 &23.91  \\
            \midrule  
            \multirow{3}{*}{\rotatebox{90}{ICSI}} & \system   &\textbf{39.66} &\textbf{10.09} &\textbf{17.53}  \\
            &  \ \ w/o \textit{utter-attn} &39.59 &9.90 &17.24  \\
            &  \ \ w/o \textit{word-attn} &38.96 &9.61 &17.40  \\
            \bottomrule 
        \end{tabular}
\caption{Ablation study for attention mechanism. \textit{utter-attn} indicates utterance-level attention and \textit{word-attn} indicates word-level attention.}
\label{tab:attention}
\end{table}

\section{Effect of the type of dialogue discourse}
Previous experiments reveal that different relations have different degrees of contribution. Thus, we filter out $N$ useless relations to see the final results (shown in Table \ref{tab:useless}). We can find that compared with using all relations, filtering some useless relations will result in performance degradation to some extent. We attribute this to the fact that although these discourse relations are of small effect, directly removing them from the meeting graph will do harm to the semantic coherence of the entire graph, which will further hurt the model performance.

\begin{table}[t]
\centering
        \begin{tabular}{clccc}
            \toprule  
            & \textbf{Model} & \textbf{R-1} & \textbf{R-2} & \textbf{R-L}  \\
            \midrule  
            \multirow{3}{*}{\rotatebox{90}{AMI}} & \system &51.42 &\textbf{20.99} &\textbf{24.89}  \\
            & filter-useless-3 &51.28  &19.68 &23.84 \\
            & filter-useless-5 & \textbf{51.44} &20.26 &24.11 \\
            \midrule  
            \multirow{3}{*}{\rotatebox{90}{ICSI}} & \system  &39.66 &\textbf{10.09} &\textbf{17.53}  \\
            & filter-useless-3 &\textbf{39.71} &9.64 &17.46 \\
            & filter-useless-5 &39.21 &9.52 &17.33 \\
            \bottomrule 
        \end{tabular}
\caption{The results of filtering out $N$ useless relations.}
\label{tab:useless}
\end{table}

\section{Case study}
Table \ref{tab:cs} shows summaries generated by different models and the visualization of utterance attention weights. The darker the color, the higher the weight.
Sentence-Gated \cite{sentgate} focuses more on the second utterance than the third utterance (shown in Table \ref{tab:cs}(a)). The second one mainly talks about ``fruit shape'' which leads to the omission of the keyword ``vegetable''.
Differently, by introducing dialogue discourse, Utterance 1 and 3 are both related to two utterances, which make them the core nodes of our graph (shown in Table \ref{tab:cs}(b)), they are discussions around the keywords ``fruit" and ``vegetable", so our model can generate a better summary that contains both keywords.

\begin{table*}[]
\centering
        \begin{tabular}{|l|l|}
            \hline
            \multicolumn{2}{|c|}{ \includegraphics[scale=0.5]{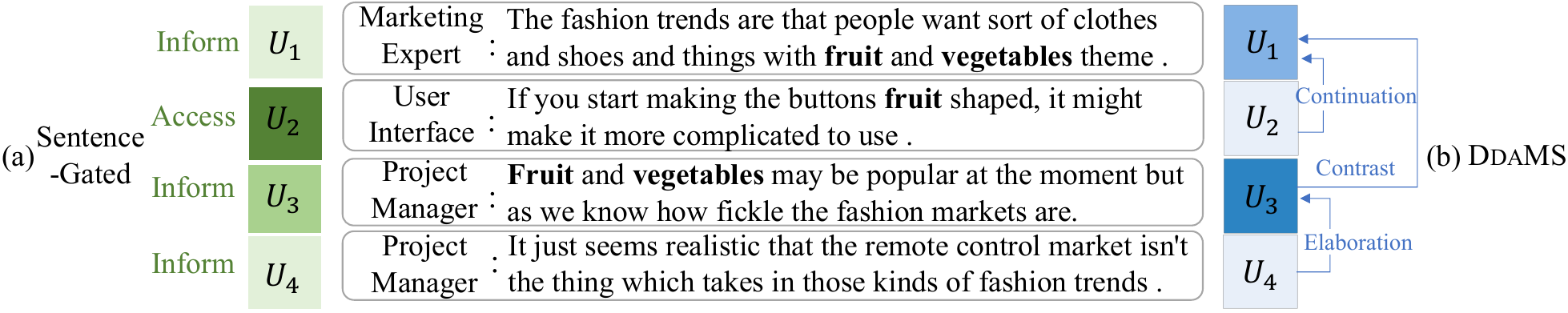}} \\
            \hline
            Ground-truth & \makecell[l]{The Marketing Expert presented trends in the remote control market and the \textbf{fruit} and \textbf{vegetable}  \\ and spongy material trends in fashion.}  \\
            \hline
            Pointer-Generator & They discussed the possibility of a \textbf{fruit or fruit and fruit}.\\
            \hline
            Sentence-Gated & \makecell[l]{The need to incorporate a \textbf{fruit} theme into the design of the remote.} \\
            \hline
            \system& \makecell[l]{The buttons will be included in a \textbf{fruit} and \textbf{vegetable} theme into the shape of the remote control.} \\
            \hline
        \end{tabular}
\caption{Example summaries generated by different models and utterance attention weights visualization of (a) Sentence-Gated (with dialogue acts) and (b) \system (with dialogue discourse relations).}
\label{tab:cs}
\end{table*}

\bibliographystyle{named}
\bibliography{ijcai21}

\end{document}